\documentclass[10pt,twocolumn,letterpaper]{article}

\usepackage{authblk}

\usepackage{cvpr}              % To produce the CAMERA-READY version
\usepackage[dvipsnames]{xcolor}

\definecolor{cvprblue}{rgb}{0.21,0.49,0.74}
\usepackage[pagebackref,breaklinks,colorlinks,citecolor=cvprblue]{hyperref}

\usepackage{multirow}  % For multi-row cells
\usepackage{array}     % For extra table formatting
\usepackage{tikz}

\usepackage{pgfplots}
\pgfplotsset{compat=1.18} % Ensure compatibility with your LaTeX version
\usepackage{pgfplotstable}
\usepackage{xcolor}
\usepackage{threeparttable}
\usepackage{listings} % Package for including code
\lstdefinestyle{pythonstyle}{
  language=Python,
  backgroundcolor=\color{gray!10}, % Light gray background
  basicstyle=\ttfamily\footnotesize, % Monospace font, small size
  keywordstyle=\color{blue}, % Keywords in blue
  commentstyle=\color{green!50!black}, % Comments in green
  stringstyle=\color{red!70!black}, % Strings in red
  numberstyle=\tiny\color{gray}, % Line numbers in gray
  numbers=left, % Add line numbers
  stepnumber=1, % Number every line
  breaklines=true, % Auto line breaking
  frame=single, % Box around the code
  captionpos=b, % Caption position below the listing
}

\def\paperID{*****} % *** Enter the Paper ID here
\def\confName{ICCV}
\def\confYear{2025}
\title{Advancing Egocentric Video Question Answering with Multimodal Large Language Models}
\author{Alkesh Patel$^*$}
\author{Vibhav Chitalia$^*$}
\author{Yinfei Yang}
\affil{Apple\vspace{-1cm}}

\begin{document}
\maketitle
\def\thefootnote{*}\footnotetext{These authors contributed equally to this work}
\begin{abstract}

Egocentric Video Question Answering (QA) requires models to handle long-horizon temporal reasoning, first-person perspectives, and specialized challenges like frequent camera movement. This paper systematically evaluates both proprietary and open-source Multimodal Large Language Models (MLLMs) on QaEgo4Dv2—a refined dataset of egocentric videos derived from QaEgo4D. Four popular MLLMs (GPT-4o, Gemini-1.5-Pro, Video-LLaVa-7B and Qwen2-VL-7B-Instruct) are assessed using zero-shot and fine-tuned approaches for both OpenQA and CloseQA settings. We introduce QaEgo4Dv2 to mitigate annotation noise in QaEgo4D, enabling more reliable comparison. Our results show that fine-tuned Video-LLaVa-7B and Qwen2-VL-7B-Instruct achieve new state-of-the-art performance, surpassing previous benchmarks by up to +2.6\% ROUGE/METEOR (for OpenQA) and +13\% accuracy (for CloseQA). We also present a thorough error analysis, indicating the model’s difficulty in spatial reasoning and fine-grained object recognition—key areas for future improvement.
\end{abstract}

\section{Introduction}
\label{sec:intro}

Egocentric video understanding—the ability to interpret long, first-person videos captured from a head-mounted camera—has major implications for augmented reality (AR), assistive AI, and personal memory retrieval. A central challenge is Egocentric Video Question Answering (QA), in which the model must parse potentially lengthy video clips to answer questions about events, objects, or actions from a first-person perspective.

Why is Egocentric Video QA hard? First, the camera often moves unpredictably, forcing models to handle significant visual blur and changing viewpoints. Second, long-term reasoning is typically required: objects may reappear after minutes, or a key event spans multiple distant time segments. Third, personalization is central; references like “Where did I put my phone?” imply the camera wearer’s viewpoint and interactions.

While Multimodal Large Language Models (MLLMs) have shown promise in video-based tasks, their performance on long-form egocentric video QA remains under-explored. Existing video QA datasets often involve short and more stable camera scenarios, limiting their applicability to realistic, first-person videos.

Despite recent progress, no prior work has systematically evaluated MLLMs on egocentric video QA benchmarks. Moreover, existing datasets, such as QaEgo4D \cite{qaego4d2022}, contain annotation inconsistencies that hinder accurate model performance assessment.

To address these limitations, we systematically investigate MLLMs on an egocentric dataset, QaEgo4Dv2, a refined version of QaEgo4D. This paper makes the following key contributions:

\begin{itemize}
\item \textbf{QaEgo4Dv2}: We clean up and refine the QaEgo4D dataset, reducing 1.2\%-3.2\% annotation errors, to eliminate inconsistent question-answer pairs and standardize answer formats, yielding more reliable annotations to assess model performance for egocentric video QA.

\item \textbf{Systematic MLLM Benchmarking}: We evaluate four state-of-the-art MLLMs—GPT-4o \cite{openai2024gpt4o}, Gemini-1.5-Pro \cite{gemini1_5Pro2024}, Video-LLaVa-7B \cite{lin2023video}, and Qwen2-VL-7B-Instruct \cite{wang2024qwen2vl} —across zero-shot and fine-tuning setups for both OpenQA and CloseQA.

\item \textbf{Empirical Study of Training Strategies}: We analyze how frame sampling strategies, supervised fine-tuning, contrastive learning, curriculum learning, and multi-task training affect QA performance.

\item \textbf{New State-of-the-Art}: Our fine-tuned Video-LLaVa-7B and Qwen2-VL-7B-Instruct model outperforms previous benchmarks, achieving up to +2.6\% improvement in OpenQA and  +13\%  improvement in CloseQA over prior methods on standard metrics.

\item \textbf{Error Analysis \& Insights}: We categorize key error modes (spatial location, object color/recognition, counting, etc.), highlighting directions for improving MLLMs in egocentric scenarios.

\end{itemize}

\section{Related Work}
\subsection{Video Question Answering}
Video QA involves answering natural language queries by analyzing temporal and spatial features from video content. Datasets like ActivityNet-QA \cite{Yu_2019} focus on human activities, requiring models to understand complex activities and interactions, while How2QA \cite{Li_2020}, built from instructional videos, emphasizes sequential processes. Datasets such as NExT-QA \cite{Xiao_2021} and Causal-VidQA \cite{Li_2022} focus on causal and temporal reasoning in video content. However, these benchmarks predominantly feature short videos, limiting their applicability to long, real-life video sequences. Recent benchmarks like Video-MME \cite{fu2024videommefirstevercomprehensiveevaluation}, MLVU \cite{zhou2024mlvucomprehensivebenchmarkmultitask}, and LongVideoBench \cite{wu2024longvideobenchbenchmarklongcontextinterleaved} include longer videos but lacks strong representation of egocentric videos, which are central to this work.

\subsection{Egocentric Video Question Answering}
Egocentric video understanding poses unique challenges, such as capturing subjective perspectives and reasoning about interactions across extended temporal sequences. Datasets like Ego4D \cite{Grauman_2022} have advanced the field by offering comprehensive benchmarks for question answering with temporal localization. For example, one of the episodic memory tasks NLQ requires localizing where the answer can be seen within the user’s past video. Building on this, QaEgo4D \cite{qaego4d2022} offers the largest egocentric video QA dataset to date, providing textual answers linked to ground-truth answer intervals and with an average video length exceeding 8 minutes. While other works, such as EM-EQA in OpenEQA \cite{OpenEQA2023}, EgoVQA \cite{EgoVQA_Fan}, EgoThink \cite{Cheng_2024}, EgoTaskQA \cite{jia2022egotaskqa}, and EgoSchema \cite{mangalam2023egoschemadiagnosticbenchmarklongform}, explore related tasks, they are limited by short video durations (less than 180 seconds), smaller dataset sizes, or questions that do not align with typical common virtual assistant use cases. More recently, EgoMemoria \cite{ye2024mmegobuildingegocentricmultimodal} includes medium (2–20 min) and long (20–60 min) videos, broadening temporal coverage. However, its automatically generated QA pairs may lack the diversity of human annotations.  Our work builds on the QaEgo4D dataset, introducing QaEgo4Dv2 to further improve annotation quality and address these limitations in long-form egocentric video question answering.

\subsection{Multimodal Large Language Models for Video QA}
MLLMs (e.g. Video-LLaVA \cite{lin2023video}, LLaVa-NeXT-Video \cite{zhang2024llavanextvideo}, Video-LLaMA \cite{zhang2023videollama}, Qwen2-VL\cite{wang2024qwen2vl}, Gemini-1.5-Pro \cite{gemini1_5Pro2024}, and GPT-4o \cite{openai2024gpt4o}),  extend traditional large language models (LLMs) by integrating vision encoders. MLLMs have been applied to various video QA benchmarks involving short videos, typically ranging from a few seconds to a couple of minutes \cite{Jang_2017, MSVDQA, Xiao_2021, Yu_2019}. More recent efforts have extended MLLMs to handle long video understanding tasks, targeting videos spanning minutes to hours \cite{li2023llamavidimageworth2, liu2024llavanext, zhang2023videollama, xu2024pllavaparameterfreellava, zhang2024longcontexttransferlanguage}. These models have been evaluated on long video question answering benchmarks \cite{fu2024videommefirstevercomprehensiveevaluation, zhou2024mlvucomprehensivebenchmarkmultitask, wu2024longvideobenchbenchmarklongcontextinterleaved}, but they do not primarily address first-person perspective leaving a gap in understanding their performance on real-world, first-person video scenarios. Our study provides a first, in-depth look at how MLLMs handle the unique challenges of egocentric QA through extensive evaluations and fine-tuning experiments.

\section{Methodology}
\label{sec:method}
In this section, we detail our approach for evaluating MLLMs on egocentric video question answering. We begin by introducing QaEgo4Dv2, an improved version of the QaEgo4D dataset, then describe the MLLMs chosen for benchmarking, and finally explain our training strategies.

\subsection{Dataset Preparation: QaEgo4Dv2}
\label{subsec:data_prep}
To benchmark MLLMs, we carefully evaluated the egocentric video QA datasets available at the time of writing. The comparison of candidate datasets is shown in Table~\ref{tab:data_comparison}. We chose QaEgo4D dataset for our experiments because it was meeting desired criteria, such as human annotated QA pairs, long video length, bigger dataset size with diversity, and types of QA pairs that require long-horizon temporal reasoning in practical applications.  While the QaEgo4D dataset was best-suited dataset for real-world utility, during preliminary evaluations, we identified annotation inconsistencies that negatively impacted model performance. To mitigate these issues, we introduce QaEgo4Dv2, a cleaned version of QaEgo4D with improved annotation accuracy.
\begin{table}[ht]
\centering
\caption{Candidate datasets considered for benchmarking Egocentric Video Question Answering task}
\label{tab:data_comparison}
\begin{tabular}{ p{1.7cm}|p{1.4cm}|p{1.1cm}|p{1cm}| p{1cm} }
\hline
\textbf{Dataset} & \textbf{Curation Method} & \textbf{Video Length} & \textbf{\# Videos} & \textbf{\# QA Pairs} \\
\hline
EgoVQA & Human & 5-10m & 16 & 580 \\
EgoTaskQA & Automatic  & 36.9s & 2,336 & 40,000 \\
EgoSchema & Human & 3m & 1,981 & 5000 \\
OpenEQA & Human & 20s & 187 & 1,636 \\
EgoMemoria & Automatic & 30s-1hr & 629 & 7,026\\
\textbf{QaEgo4D} & \textbf{Human} & \textbf{8.2m} & \textbf{1,325} & \textbf{14,507}  \\
\hline
\end{tabular}
%\vspace{-3mm}
\end{table}

We systematically removed noisy instances from QaEgo4D based on the following criteria:
\begin{itemize}
\item \textbf{Fixed question template errors}: Instances where ``l'' was mistakenly used instead of ``I'' (e.g., ``Where did l put foil paper?'' → ``Where did I put foil paper?''). These errors affected 1.2–4.3\% of training, validation, and testing instances.
\item \textbf{Corrected answer mismatches}: Responses that were semantically inconsistent with the question type (e.g., ``Where was my remote control in the living room?'' → “Yes” instead of a location). These question-answer mismatches impacted 1-2.6\% training, validation, and testing instances. 
\item \textbf{Standardized answer formats}: Ensured answers were concise and aligned with model output expectations.
\end{itemize}

We release script to generate QaEgo4Dv2 dataset in supplementary materials. Appendix~\ref{sec:appendix_d} shows more examples of noise in dataset. We do not modify minor grammatical or spelling errors, nor do we adjust awkward phrases as we observe that MLLMs are robust to them. Additionally, we retain the ground truth answer intervals as they were from the original dataset, as rectifying them would require significant manual annotations, given that automatic techniques proved unreliable. We leave this manual verification as future work. 

After cleaning, QaEgo4Dv2 contains:
\begin{itemize}
\item 10,406 training instances (↓3.2\% from 10,746 due to removed errors).
\item 1,880 validation instances (↓1.6\% from 1,911).
\item 1,826 test instances (↓1.3\% from 1,850).
\end{itemize}

Table~\ref{tab:data_cleaning} in Section~\ref{sec:results} highlights the impact of dataset refinement on model performance before-and-after corrections.

% \begin{table}[ht]
% \centering
% \resizebox{\columnwidth}{!}{%
% \begin{tabular}{|c|c|c|c|c|c|c|}
% \hline
% \multirow{2}{*}{\textbf{Model}} & \multirow{2}{*}{\textbf{Trained On}} & \multicolumn{5}{c|}{\textbf{Eval on QAEgo4D original test set}} \\ \cline{3-7}
%  &  & Similarity & GPT-4 Acc & GPT-4 Score & Gemini Acc & Gemini Score\\ \cline{1-7}
% \multirow{3}{*}{Video-LLaVa} & Original & 0.5657 & 0.3019 & 1.6401 & 0.3924 & 1.8915 \\ \cline{2-7}
%  & Cleaned & 0.5799 & 0.3165 & 1.7619 & 0.412 & 1.9913 \\ \cline{2-7}

%  & \boldsymbol{\Delta} & 1.42\% & 1.46\% & 12.18\% & 1.96\% & 9.98\% \\ \hline
% \end{tabular}%
% }

% \caption{Result for Video-LLaVa trained on original and cleaned data. Evaluated on original test set}
% \end{table}

%-------------------------------------------------------------------------

%-------------------------------------------------------------------------
\subsection{Model Selection: Evaluating Multimodal LLMs}
We benchmark four MLLMs that accept video as input:
\begin{itemize}
	\item	\textbf{GPT-4o} \cite{openai2024gpt4o} (proprietary)
	\item	\textbf{Gemini-1.5-Pro} \cite{gemini1_5Pro2024} (proprietary)
	\item	\textbf{Video-LLaVa-7B} \cite{zhu2023languagebind} (open-source)
	\item	\textbf{Qwen2-VL-7B-Instruct} \cite{wang2024qwen2vl} (open-source)
\end{itemize}

GPT-4o and Gemini-1.5-Pro are widely known for their strong language and vision capabilities, though their fine-tuning is either limited or expensive. By contrast, Video-LLaVa and Qwen2-VL are publicly available in Hugging Face \cite{huggingface} for full fine-tuning. Below we describe our approach to testing each.
%\vspace{-3mm}
\subsubsection{Prompt Design for Pre-trained MLLMs}
%\vspace{-2mm}
  We observed that off-the-shelf pre-trained MLLMs either produce overly verbose outputs (e.g., GPT-4o, Gemini-1.5-Pro) or generate nonsensical outputs (e.g., Video-LLaVa) when provided only with a question and video. QaEgo4D ground truth annotations are short phrases, e.g., for \textit{``Where did I drop the cutter?''}, the expected answer is \textit{``on the table''} rather than a full sentence like \textit{``You dropped the cutter on the table''}. To ensure model outputs align with the desired format and mitigate nonsensical responses, we meticulously designed prompts to evaluate its inherent capabilities. The specific zero-shot prompts used for OpenQA and CloseQA in our experiments are detailed in Appendix~\ref{sec:appendix_a}. 
% \begin{table}[ht]
% \centering
% \caption{Models selected for experiments}
% \label{tab:model_selection}
% \begin{tabular}{ |p{2cm}|p{1.5cm}|p{1.5cm}|p{1.5cm}|} 
% \hline
% \textbf{Model} & \textbf{Type} & \textbf{Video Input Handling} & \textbf{Fine-Tuning Capability}
% \\
% \hline
% GPT-4o & Proprietary & Direct vision-text fusion & x Not trainable  \\
% Gemini-1.5-Pro & Proprietary  & Optimized for multimodal & x Not trainable \\
% Video-LLaVa & Open-source & CLIP-based video encoding & \checkmark Trainable \\
% LLaVa-NeXT-Video & Open-source & Advanced ViT integration & \checkmark Trainable  \\
% \hline
% \end{tabular}
% \end{table}
% \vspace{-2mm}
\subsubsection{Fine-Tuning Strategies for MLLMs}
For open-source models (Video-LLaVa, Qwen2-VL), we explore four fine-tuning methods. For each method we chose, we present our \textit{hypothesis} and \textit{approach}.

\begin{itemize}
\setlength{\itemsep}{2pt} 
\item \textbf{Standard Supervised Fine-Tuning (SFT)}

\underline{\textit{Hypothesis}}: Directly training on human-annotated (video, question → answer) data should better align the model with egocentric QA demands.

\underline{\textit{Approach}}: We randomly sample from QaEgo4Dv2 in each training batch and optimize standard supervised loss.

\item \textbf{Contrastive Fine-Tuning}

\underline{\textit{Hypothesis}}: The model should perform better when given an oracle clip (ground truth time interval) than when given the full video. However, prior work \cite{Di_2024} has shown that this performance gap is small, indicating suboptimal temporal localization. Training with contrastive pairs should improve the model’s ability to distinguish between relevant and irrelevant segments.

\underline{\textit{Approach}}: For each QA pair, we sample positive clips (ground-truth intervals) and negative clips (no overlap with ground-truth). We add a contrastive loss to encourage the model to learn discriminative embeddings.
Note: We generate negative intervals randomly from the same video—ensuring no overlap with ground-truth segments—as well as from different videos.

\item \textbf{Curriculum Learning}
\label{sec:curriculum}

\underline{\textit{Hypothesis}}: Gradually exposing the model to simpler QA tasks (e.g., object recognition) before more complex ones (e.g., event or spatial queries) might improve final performance \cite{BengioCurriculum}.

\underline{\textit{Approach}}: We categorize each training instance as easy, medium, or hard based on its label loss computed using a pre-trained MLLM (e.g., Video-LLaVa and Qwen2-VL). Instances with lower loss are considered easier. During training, we introduce instances in order of increasing difficulty, progressing from easier to harder examples.

\item \textbf{Multi-Task Fine-Tuning}

\underline{\textit{Hypothesis}}: Joint optimization for both OpenQA and CloseQA fosters more robust language-video alignment and thereby improving generalization.

\underline{\textit{Approach}}: We set a multi-task loss on OpenQA (short textual answer) and CloseQA (multiple-choice) samples, weighting them equally.
\end{itemize}

\section{Experimental Setup}

%\subsection{Experimental Setup}

\subsection{Datasets}
We use a refined version of the original QaEgo4D dataset i.e., QaEgo4Dv2, which includes annotations for both OpenQA and CloseQA tasks for all our experiments, except for comparison with prior work (see Section~\ref{sec:prior_work}). 

\subsection{Evaluation Metrics}
\textbf{OpenQA}: For OpenQA, we find that traditional metrics like ROUGE and METEOR, while providing a high-level overview of answer quality, are suboptimal for this task. Instead, we leverage state-of-the-art proprietary models, GPT-4o and Gemini-1.5-Flash \cite{gemini1_5Pro2024}, as judges to produce quality estimates that align closely with human evaluations. The use of LLMs-as-judges \cite{Zheng2023JudgingLW} is common in text-based evaluations within the community. However, unlike previous work \cite{xu2023retrievalbasedvideolanguagemodel}, which evaluates generated and ground truth answers solely in the context of the question, our approach incorporates video context into the grading process. This approach results in more reliable evaluation metrics, albeit with higher API usage costs. Nevertheless, this approach remains significantly more efficient and less time-consuming than human judgments. 

We prompt GPT-4o and Gemini-1.5-Flash to compare the predicted and ground truth answers within the context of the input video (32 evenly sampled frames from video), determine their semantic equivalence, and assign a similarity score from 0 to 5. Accuracy (GPT/Gemini Acc) and average scores (GPT/Gemini Score) across the test set are reported. Additionally, we track performance using an average similarity score (Sim.) obtained from a sentence transformer \cite{Reimers2019SentenceBERTSE}, which measures text-based similarity independent of video content.\\
\\
\textbf{CloseQA}: For CloseQA, the evaluation metric is accuracy (Acc) in selecting the correct answer from a set of multiple choices. To mitigate biases arising from potential learning of answer-choice order during pre-training or fine-tuning, we evaluate the model across five random seed configurations. Each configuration randomizes the order of answer choices, and the final accuracy is reported as the average across these runs.\\ 
\\
The prompts used for automatic evaluation are detailed in Appendix~\ref{sec:appendix_b}.
\subsection{Video Input Variants}
We consider multiple ways to feed video frames to MLLMs:
\begin{itemize}
\item \textbf{\textit{Oracle clip}}: Evenly sampled frames from ground truth interval to measure the upper bound on model performance.
\item \textbf{\textit{No-video (text-only)}}: The model sees only the question text, without video input to ablate model's language-based understanding.
\item \textbf{\textit{Full video}}: Evenly sampled frames from the entire video.
\item \textbf{\textit{Full video-GT}}: Evenly sampled frames from the entire video, excluding the ground truth (GT) interval to test whether the model truly localize relevant segments or guesses from context.
\end{itemize}

For all experiments involving video inputs—except for Video-LLaVa, which supports only 8 frames—we use 32 evenly sampled frames in their respective video input variants. To analyze the impact of the number of frames (see Table~\ref{tab:imframes}), we also experiment with 8, 16, and 24 frames, evenly sampled across their respective video input variants. 

To assess the model’s reliance on visual data, we also compare these setups with and without textual fine-tuning (see Table~\ref{tab:imvisual}). All fine-tuned models, except those in the no-video (text-only) setup, are trained on the full videos from the QaEgo4Dv2 dataset.

\subsection{Implementation Details}
\textbf{Training}: We use AdamW optimizer with 3-e6 learning rate and no weight decay.  We use cosine scheduler with 1000 warmup steps.\\ \textbf{Batch Size}: The effective batch size for all our experiments is 128.\\
\textbf{Resolution}: We use a resolution of 336×336 for GPT-4o, Gemini-1.5-Pro, and Qwen2-VL, except for Video-LLaVa, which uses 224×224. Ablations in Appendix ~\ref{sec:appendix_c} show that higher resolution does not always improve performance.

\subsection{Research Questions}
Based on our evaluation setup, we address the following research questions (RQ):
\begin{itemize}
\item \textbf{RQ1}: How do proprietary vs. open-source MLLMs perform on long-form egocentric video QA tasks?
\item \textbf{RQ2}: What is the impact of fine-tuning on MLLM performance?
\item \textbf{RQ3}: How does the number of frames affect video understanding capabilities?
\item \textbf{RQ4}: What are the primary failure cases in long-form egocentric video QA?
\end{itemize}

These research questions guide our quantitative and qualitative analyses in Section~\ref{sec:results}.

\section{Results and Discussion}
\label{sec:results}

% Wide table that spans both columns
\begin{table*}[ht]
\centering
\caption{Experiments with various MLLMs on QaEgo4Dv2 dataset for OpenQA and CloseQA setups. The best results are \textbf{bold}, the second-best results are \underline{underlined}, and upperbound results are \textbf{bold} \underline{underlined}. This convention is followed in all subsequent tables.}
\label{tab:mllm}
\begin{tabular}{p{2.3cm}|p{3.2cm}|p{1.5cm}|p{0.8cm}|p{0.9cm}|p{0.9cm}|p{1cm}|p{1cm}| p{2cm}}
\hline
\multirow{3}{=}{Model} & \multirow{3}{=}{Training Method} &  \multirow{3}{=}{Eval Input} & \multicolumn{5}{c|}{OpenQA} & CloseQA\\
&  &   & \multirow{2}{=}{Sim} & GPT Acc & GPT Score & Gemini Acc & Gemini Score & \multirow{2}{=}{Acc}\\
\hline
 GPT-4o  & 0-shot    & oracle clip   &  49.37   & \underline{\textbf{42.53}}   & \underline{\textbf{2.25}} & 59.39 & 2.83  & $45.65 \pm 8.6$ \\
Gemini-1.5-Pro  & 0-shot  & oracle clip & \underline{\textbf{54.27}}   & 41.69 & 2.22 & \underline{\textbf{62.83}} & \underline{\textbf{2.99}} & $\underline{\textbf{65.56}} \pm 2.1$ \\
\hline
\hline
GPT-4o  & 0-shot  & full video & 39.94 & 21.77 & 1.18 & 30.53 & 1.46 & $26.28 \pm 8.1$\\
Gemini-1.5-Pro  & 0-shot  & full video   &  48.46   & 29.10   & 1.56 & \textbf{43.87} & \textbf{2.07}  & $50.76 \pm 1.4$ \\
Video-LLaVa    & 0-shot  & full video & 47.38  & 22.96 & 1.27 & 30.03 & 1.42 & $37.40 \pm 12$\\
Qwen2-VL    & 0-shot  & full video & 52.95  & 29.98 & 1.62 & 37.69 & 1.80 & $40.00 \pm 5.2$\\
Video-LLaVa   & sft \footnotesize{(supervised fine-tuning)} & full video & \textbf{58.06} & \textbf{32.11} & \textbf{1.74} & \underline{40.75} & \underline{1.95} & $\underline{51.88} \pm 5.0$\\

Qwen2-VL & sft & full video & \underline{57.43} & \underline{31.40} & \underline{1.72} & 39.50 & 1.92 & $\textbf{55.04} \pm 4.2$ \\
\hline
\end{tabular}
%\vspace{-3mm}
\end{table*}

\subsection{Performance Analysis of Multimodal Large Language Models}
%\(\sim36\%\)
Evaluating proprietary and open-source MLLMs on QaEgo4Dv2 dataset (see Table~\ref{tab:mllm}) revealed key insights into their capabilities for egocentric video QA:
\begin{itemize}
\item When evaluated the upper-bound performance using GPT-4o and Gemini-1.5-Pro by providing the ground-truth answer clip, the results indicate that GPT-4o achieves 43\% GPT accuracy and 59\% Gemini accuracy, while Gemini-1.5-Pro attains 42\% GPT accuracy and 63\% Gemini accuracy. Both models remain below 70\%, highlighting the inherent difficulty of the QaEgo4Dv2 dataset and the egocentric video question answering task.

\item When evaluated on full videos, representing real-world scenarios, the performance of both proprietary models drops significantly (up to 30-48\%), indicating that understanding long egocentric videos remains a challenge. 

\item Fine-tuning publicly available MLLMs, such as Video-LLaVA and Qwen2-VL, on the QaEgo4Dv2 dataset yields competitive performance with respect to proprietary models. While Gemini-1.5-Pro maintains a lead in zero-shot performance, fine-tuned public models outperform proprietary models by a significant margin in CloseQA setup and show competitive results in OpenQA setup.

Overall, the findings in Table~\ref{tab:mllm} underscore the promise of using MLLMs for egocentric video question answering. 
\end{itemize}

\begin{table*}[ht]
\centering
\caption{Experiments to assess the impact of visual features on QaEgo4Dv2 dataset for OpenQA and CloseQA setups.}
\label{tab:imvisual}
\begin{tabular}{p{2.3cm}|p{2.5cm}|p{2cm}|p{0.8cm}|p{0.9cm}|p{0.9cm}|p{1cm}|p{1cm}| p{2cm}}
\hline
\multirow{3}{=}{Model} & \multirow{3}{=}{Training Method} &  \multirow{3}{=}{Eval Input} & \multicolumn{5}{c|}{OpenQA} & CloseQA\\
&  &   & \multirow{2}{=}{Sim} & GPT Acc & GPT Score & Gemini Acc & Gemini Score & \multirow{2}{=}{Acc}\\
\hline
\multirow{3}{=}{GPT-4o}   & 0-shot  & text-only & 24.41  & 0.27 & 0.04 & 0.44 & 0.03 & $0.8 \pm 0.3$\\
& 0-shot & full video-GT & 39.00 & 21.77 & 1.18 & 30.53 & 1.46 & $23.34 \pm 7.0$\\
& 0-shot & full video & \textbf{39.94} & \textbf{22.96} & \textbf{1.25} & \textbf{32.32} & \textbf{1.55} & $\textbf{26.28} \pm 8.1$\\
\hline
\multirow{3}{=}{Gemini-1.5-Pro}   & 0-shot  & text-only & 24.44  & 0.27 & 0.04 & 0.49 & 0.03 & $0.2 \pm 0.2$\\
& 0-shot & full video-GT & 48.15 & \textbf{29.45} & \textbf{1.57} & 42.50 & 2.01 & $49.20 \pm 1.9$\\
& 0-shot & full video & \textbf{48.46} & 29.10 & 1.56 & \textbf{43.87} & \textbf{2.08} & $\textbf{50.76} \pm 1.4$\\
\hline
\multirow{6}{=}{Video-LLava}  & 0-shot & text-only & 41.40 & 12.64 & 0.95 & 14.44 & 0.94 & $9.5 \pm 15$\\
& 0-shot & full video-GT & \textbf{47.77} & \textbf{23.97} & \textbf{1.31} & \textbf{30.66} & \textbf{1.45} & $36.84 \pm 11$\\
& 0-shot  & full video & 47.38  & 22.96 & 1.27 & 30.03 & 1.42 & $\textbf{37.40} \pm 12$\\
\cline{2-9}
& sft (text-only) & text only & 53.79 & 23.35 & 1.63 & 24.72 & 1.58 & $43.40 \pm 0.9$\\
& sft  & full video-GT &  58.02 & 31.84 & \textbf{1.75} & \textbf{40.86} & \textbf{1.95} & $51.20 \pm 5.4$\\
& sft & full video & \textbf{58.06} & \textbf{32.11} & 1.74 & 40.75 & \textbf{1.95} & $\textbf{51.88} \pm 5.0$ \\ 
\hline
\multirow{6}{=}{Qwen2-VL} & 0-shot  & text-only & 26.09 & 1.91 & 0.13 & 2.19 & 0.21 & $3.5 \pm 1.3$ \\
& 0-shot  & full video-GT & 50.24 & 28.35 & 1.52 & 34.96 & 1.69 & $39.68 \pm 5.6$ \\
& 0-shot  & full video & \textbf{52.94}  & \textbf{29.98} & \textbf{1.62} & \textbf{37.69} & \textbf{1.80} & $\textbf{40.00} \pm 5.2$\\
\cline{2-9}
& sft (text-only)  & text only & 53.79 & 23.35 & 1.63 & 24.72 & 1.58 & $42.28 \pm 1.4$\\
& sft  & full video-GT & 57.02 & 30.96 & 1.69 & 39.17 & 1.90 & $54.72 \pm 4.2$\\
& sft & full video & \textbf{57.43} & \textbf{31.40} & \textbf{1.72} & \textbf{39.50} & \textbf{1.92} & $\textbf{55.04} \pm 4.2$ \\
\hline
\end{tabular}
\end{table*}

\subsection{Impact of visual features}
Analyzing the role of video input (see Table~\ref{tab:imvisual}) in MLLM performance provided the following insights:
\begin{itemize}
\item Video input boost performance in both OpenQA (e.g., higher GPT accuracy/score) and CloseQA accuracy, emphasizing importance of visual features.

\item A text-only baseline can achieve non-zero accuracy, suggesting language priors in MLLMs (e.g., Video-LLaVa and Qwen2-VL) or repeated question formats.  For example, the question \textit{``Where did I first see the fridge?''} is likely to be answered correctly as \textit{``kitchen''}, even without visual input. 

\item Supervised text fine-tuning reduces the gap between text-only and video input performance, further emphasizing predominance of text bias.  GPT-4o and Gemini-1.5-Pro effectively addresses this bias, as evidenced by their text-only zero-shot results, which exhibit the lowest OpenQA accuracy and nearly zero CloseQA accuracy. 

\item Excluding the ground truth region (full video-GT) has little impact— slightly reducing the performance— showing MLLMs use both relevant intervals and context cues throughout the video to predict the correct answer.
\end{itemize}

\begin{table*}[ht]
\centering
\caption{Fine-tuning experiments on QaEgo4Dv2 dataset for OpenQA and CloseQA tasks}
\label{tab:imfunetuning}
\begin{tabular}{p{2.3cm}|p{2cm}|p{1.5cm}|p{0.8cm}|p{0.9cm}|p{0.9cm}|p{1cm}|p{1cm}| p{2cm}}
\hline
\multirow{3}{=}{Model} & \multirow{3}{=}{Training Method} &  \multirow{3}{=}{Eval Input} & \multicolumn{5}{c|}{OpenQA} & CloseQA\\
&  &   & \multirow{2}{=}{Sim.} & GPT Acc & GPT Score & Gemini Acc & Gemini Score & \multirow{2}{=}{Acc}\\
\hline
\multirow{4}{=}{Video-LLaVa }  & sft & full video & 58.06 & 32.11 & 1.74 & \underline{40.75} & \underline{1.95} & $\textbf{51.88}  \pm 5.0$ \\
& contrastive  & full video & \textbf{58.65} & \textbf{33.43} & \textbf{1.83} & \textbf{42.12} & \textbf{2.02} & $47.84 \pm 2.2$ \\
& curriculum & full video & \underline{58.09} & \underline{32.49} & \underline{1.76} & 39.82 & 1.93 & $\underline{50.64} \pm 8.5$\\
& multi-task  & full video & 57.79 & 32.09 & 1.73 & 39.72 & 1.91 & $50.00 \pm 6.5$\\
\hline
\multirow{4}{=}{Qwen2-VL } & sft & full video & \underline{57.43} & \underline{31.40} & \underline{1.72} & \underline{39.50} & \underline{1.92} & $\textbf{55.04} \pm 7.7$ \\
& contrastive  & full video & 57.10 & 31.27 & 1.70 & 38.68 & 1.87 & $50.04 \pm 7.7$\\
& curriculum & full video & \textbf{57.61} & \textbf{31.47} & \textbf{1.74} & \textbf{40.86} & \textbf{1.97} & $52.36 \pm 6.8$\\
& multi-task  & full video & 56.94 & 30.63 & 1.66 & 38.94 & 1.87 & $\underline{54.44} \pm 8.2$\\
\hline
\end{tabular}
\end{table*}

\subsection{Impact of fine-tuning strategies}
Observations from Table~\ref{tab:imfunetuning} on fine-tuning techniques are summarized below:
\begin{itemize}
\item All four fine-tuning strategies (sft, contrastive, curriculum, and multi-task) significantly outperform zero-shot pre-trained performance (up to 40\% improvements), standard supervised fine-tuning often yielding the consistent performance across MLLMs and metrics.

\item Contrastive learning helps Video-LLaVa slightly but does not consistently improve on Qwen2-VL, perhaps due to overlapping design philosophies (e.g., use of cross-attention compression that reduces redundancy while maintaining positional cues).

\item Curriculum learning shows improvements in Qwen2-VL, suggesting its efficacy. However, a more nuanced scheduling or difficulty estimation is needed. Experiments with adaptive curriculum \cite{9737531} strategies are left as future work.

\item Multi-task fine-tuning did not outperform standard supervised fine-tuning, suggesting that combining OpenQA and CloseQA led to suboptimal representations compared to task-specific fine-tuning.

\item None of the fine-tuning strategies—contrastive, curriculum, or multi-task—improved the results for the CloseQA setup. Nevertheless, all results were consistently better than those of proprietary models like GPT-4o and previous methods such as GroundVQA.
\end{itemize}

% \subsection{Impact of cleaned data}
% As discussed in the Section 2, QaEgo4D dataset had about 3\% noise in its training split and 1.2\% noise in validation split.  Table 3 shows the effect of cleaning the training data.  We show the results for both original (test set with 1\% noise) and cleaned test set for fair comparison with other work.  We notice that having higher quality training set does matter. We observe x\% improvement after fine-tuning MLLM with cleaned QaEgo4Dv2 dataset.  For brevity, we show GPT4-Accuracy and Gemini-Accuracy here. For more details, please refer Appendix 3.  We also show in Appendix 3 that adding more data doesn't help as far as fine-tuning MLLM is concerned. Smaller dataset with high quality makes real difference.

% \begin{table}[ht]
% \centering
% \caption{Ablation on cleaned QaEgo4D and QaEgo4Dv2 datasets when fine-tuning Video-LLaVa on OpenQA task}
% \begin{tabular}{ p{1.7cm}|p{1cm}|p{1cm}| p{1cm}|p{1cm} } 
% \hline
% Train Data  & \multicolumn{2}{c|}{QaEgo4D Test} & \multicolumn{2}{c}{QaEgo4Dv2 Test}\\
% & GPT-A & Gem-A & GPT-A & Gem-A \\
% \hline
% QaEgo4D & 31.18 & \textbf{40.71} & 31.26 & \textbf{40.67}   \\ 
% QaEgo4Dv2 & \textbf{31.29} & 40.67 & \textbf{31.45} & 40.53 \\ 
% \hline
% \end{tabular}
% \end{table}

\begin{table*}[ht]
\centering
\caption{Experiments with number of frames given to MLLMs on QaEgo4Dv2 dataset}
\label{tab:imframes}
\begin{tabular}{p{2.3cm}|p{2.5cm}|p{1.2cm}|p{0.8cm}|p{0.9cm}|p{0.9cm}|p{1cm}|p{1cm}| p{1.6cm}}
\hline
\multirow{3}{=}{Model} & \multirow{3}{=}{Training Method} & \multirow{3}{=}{Frames} & \multicolumn{5}{c|}{OpenQA} & CloseQA\\
&  & & \multirow{2}{=}{Sim.} & GPT Acc & GPT Score & Gemini Acc & Gemini Score & \multirow{2}{=}{Acc}\\
\hline
\multirow{4}{=}{GPT-4o}  & \multirow{4}{=}{0-shot} & 8 &  38.06   & 19.96 & 1.07 & 27.16 & 1.30 & $23.53 \pm 5.3$ \\
&  & 16    &   39.55 & 22.76 & 1.22 & 31.24 & 1.49 & $24.58 \pm 6.1$ \\
&  & 24    & \textbf{40.21}   & \textbf{23.53} & \textbf{1.27} & \textbf{32.66}  & \textbf{1.56} & $24.45 \pm 6.7$ \\
&  & 32 & 39.94   & 22.96   & 1.25 & 32.32 & 1.55  & $\textbf{26.28} \pm 8.1$ \\
\hline
\multirow{4}{=}{Gemini-1.5}  & \multirow{4}{=}{0-shot} & 8  & 48.10   & 26.72 & 1.43 & 37.75 & 1.81 & $49.08 \pm 1.1$ \\
& & 16  & 48.23  & 27.43 & 1.51 & 41.65 & 1.97 & $51.20 \pm 1.7$ \\
& & 24  & 48.40  & \textbf{29.10} & \textbf{1.57} & 41.85 & 1.99 & $\textbf{50.76} \pm 1.4$ \\
& & 32  & \textbf{48.46}   & \textbf{29.10} & 1.56 & \textbf{43.87} & \textbf{2.08} & $\textbf{50.76} \pm 1.4$ \\
\hline
\multirow{4}{=}{Qwen2-VL}  & \multirow{4}{=}{sft} & 8  & 57.00   & 30.07 & 1.67 & 37.91 & 1.86 & $54.40 \pm 5.1$ \\
& & 16  & 57.08  & 30.60 & 1.69 & 39.17 & 1.90 & $53.88 \pm 5.0$ \\
& & 24  & 57.40  & \textbf{31.76} & \textbf{1.74} & \textbf{39.68} & \textbf{1.92} & $54.16 \pm 4.8$ \\

& & 32  & \textbf{57.43} & 31.40 & 1.72 & 39.50 & \textbf{1.92} & $\textbf{55.04} \pm 4.2$ \\
\hline
\end{tabular}
\end{table*}

\subsection{Impact of number of frames}
\label{sec:frames_impact}
 We make two main observations in Table~\ref{tab:imframes} when assessing the impact of number of frames:
 \begin{itemize}
\item Increasing frames from 8 to 16 or 24 generally boosts performance by 4–14\%, but returns diminish beyond 24 frames. 
 
\item Qwen2-VL with 32 frames obtains the highest CloseQA accuracy, at the cost of greater computational load. Future models may need more efficient strategies for truly long video contexts (e.g., 8+ minutes).
\end{itemize}

% \subsection{Impact of image resolution}
% Our intuition was that by increasing the resolution of the image, model will be able to see the smaller objects better and subsequently perform better on at least on questions related to object detection (e.g. What did I X) and color (e.g. What color is X). However, increasing the resolution of the image didn't make significant difference in overall results. We believe reason for this behavior could be that MLLMs used in the experiments are already leveraging the required resolution and increasing it may not be necessary. Another reason (as we observed earlier) could be the dominance of text modality in MLLM that dilutes the potential gains anticipated from higher resolution images.

\subsection{Impact of dataset refinements}
Table~\ref{tab:data_cleaning} shows impact of systematically removing noise from QaEgo4D dataset as discussed in Section~\ref{subsec:data_prep}. We notice 3-4\% gains in GPT/Gemini accuracy and 2-3\% gains in GPT/Gemini scores when fine-tuning Video-LLaVA model with and without cleaned data, emphasizing effectiveness of QaEgo4Dv2 dataset. Although some residual noise may remain (e.g., ambiguous questions), the refined dataset significantly reduces QA mismatches.

\begin{table*}[ht]
\centering
\caption{Impact of dataset refinements. Video-LLaVa is trained on original and cleaned data. Evaluated on original test set}
\label{tab:data_cleaning}
\begin{tabular}{c|c|c|c|c|c|c}
\hline
\multirow{2}{*}{\textbf{Model}} & \multirow{2}{*}{\textbf{Trained On}} & \multicolumn{5}{c}{\textbf{OpenQA - Evaluated On QaEgo4D}} \\ %\cline{3-7}
 &  & Sim. & GPT Acc & GPT Score & Gemini Acc & Gemini Score\\ \cline{1-7}
\multirow{2}{*}{Video-LLaVa} & QaEgo4D & 56.83 & 30.55 & 1.68 & 39.08 & 1.88 \\ 
 & QaEgo4Dv2 & 57.65 \textbf{(+1.4\%)}& 31.68 \textbf{(+3.7\%)}  & 1.72 \textbf{(+2.3\%)} & 40.54 \textbf{(+3.7\%)} & 1.93 \textbf{(+2.7\%)} \\ 
% \hline
% \multirow{2}{*}{Qwen2-VL} & QaEgo4D & 56.73 & 31.62 & 1.71 & 39.62 & 1.92 \\ 
% & QaEgo4Dv2 & 57.43 \textbf{(+1.2\%)}  & 31.40 \textbf{(-0.01\%)}  & 1.72 \textbf{(+1.1\%)} & 39.50 \textbf{(+1\%)} & 1.92 \textbf{(+1\%)}\\ 

% & $\boldsymbol{\Delta}$ & \textbf{+1.42\%} & \textbf{+1.46\%} & \textbf{+12.18\%} & \textbf{+1.96\%} & \textbf{+9.98\%} \\ 
 \hline
\end{tabular}
\end{table*}

\subsection{Comparison with Prior Work}
\label{sec:prior_work}
We report results on the original (uncleaned) QaEgo4D dataset, using the same baseline models and evaluation metrics from \cite{Di_2024} for fair comparison.
\\
\textbf{GroundVQA} was trained for both OpenQA and CloseQA tasks simultaneously.
\\
\textbf{GroundVQA+} extends GroundVQA by including VLG (video language grounding task).
\\
\textbf{GroundVQA++} integrates EgoTimeQA \cite{Di_2024} data on top of GroundVQA.
\\
\textbf{GroundVQA+++} integrates EgoTimeQA data on top of GroundVQA+.
\\
Table~\ref{tab:sota} provides a comparison of the best-performing fine-tuning recipes for the Video-LLaVa and Qwen2-VL models against previous state-of-the-art methods:
\begin{itemize}
\item Contrastive fine-tuning on Video-LLaVa outperforms GroundVQA and its variants by up to +2.6\% in OpenQA metrics, while standard supervised fine-tuning on Qwen2-VL achieves a +13\% improvement in CloseQA accuracy, setting a new state-of-the-art on the QaEgo4D benchmark.

\item More crucially, fine-tuned Video-LLaVa and Qwen2-VL accomplish this without adding extra tasks or datasets (e.g., no specialized VLG or EgoTimeQA\cite{Di_2024}  data). Fine-tuning a general-purpose MLLM on a QaEgo4D dataset indicates that large-scale models can excel at specialized tasks given enough domain alignment.
\end{itemize}

\begin{table}[ht]
\centering
\caption{Comparison of state-of-the-art methods with fine-tuned MLLMs (R = ROUGE, M = METEOR) }
\label{tab:sota}
\begin{tabular}{ p{2.6cm}|p{0.6cm}|p{0.6cm}|p{0.6cm}|p{1.6cm} } 
\hline
Method & \multicolumn{3}{|c|}{OpenQA} & CloseQA \\
& Sim. & R & M & Acc \\
\hline
% MFAS & - & 28.2 & 18.9 & -\\
GroundVQA & 54.8 & 27.7 & 18.7 & $39.5\pm 0.5$ \\ 
$\text{GroundVQA}^{+}$ & 55.6 & 29.0 & 19.8 & $40.8 \pm 1.0$ \\ 
$\text{GroundVQA}^{++}$ & 56.1 & 28.8 & 20.1 & $47.2 \pm 0.5$ \\ 
$\text{GroundVQA}^{+++}$ & 57.7 & 30.2 & 21.2 & $48.7 \pm 0.4$ \\
% \hline
% SimpleVQA & 54.9 & 28.0 & 19.0 & $41.3 \pm 0.4$ \\ 
% SimpleVQA+ & 56.1 & 28.8 & 20.2 & $47.1 \pm 0.3$\\ 
\hline
$\text{Video-LLaVa}^{*}$ & \textbf{58.3} & \textbf{31.0} & \textbf{21.4} & $47.8 \pm 2.2$\\
$\text{Qwen2-VL}^{**}$ & 56.9 & 30.1 & 20.6 & $\textbf{55.0} \pm 4.2$\\ 
% Our Recipe++ & 59.0 & 31.1 & 21.7 & 47.3 \\ 
\hline
\end{tabular}
\footnotesize{* with contrastive fine-tuning, ** with standard supervised fine-tuning.}
%\vspace{-3mm}
\end{table}

\subsection{Qualitative Analysis}
\begin{figure*}[ht]
    \centering
    \caption{Qualitative analysis of various MLLMs}
    \includegraphics[width=1.0\textwidth]{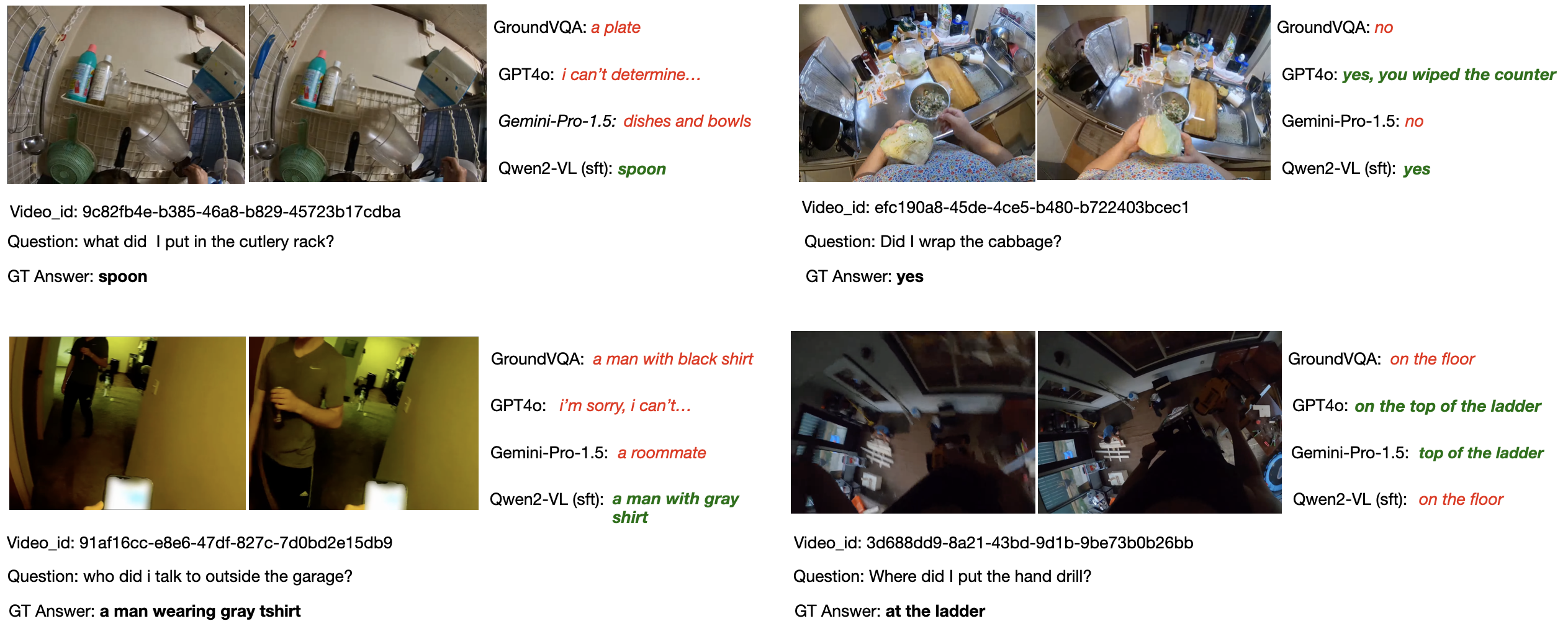} % Replace with your image file
    \label{fig:quality}
\vspace{-4mm}
\end{figure*}

For qualitative analysis, we compared answers generated by GroundVQA, GPT-4o, Gemini-1.5-Pro, and fine-tuned Qwen2-VL with the ground truth answers from the QaEgo4Dv2 dataset. We computed the semantic similarity of generated answer with the ground truth using sentence transformer \cite{Reimers2019SentenceBERTSE}. A model was considered the winner if its similarity score was at least 0.8 and higher than the others. The 0.8 threshold is arbitrary, intended to provide a general sense of performance. Out of 259 instances that met this threshold, we observed that fine-tuned Qwen2-VL won 110 times (42\%), GPT-4o won 33 times (13\%), Gemini-1.5-Pro won 59 times (23\%), and GroundVQA won 57 times (22\%). Figure~\ref{fig:quality} shows examples illustrating each model's strengths and weaknesses.

% Insert a figure

\subsection{Error Analysis}
We investigated fine-tuned Qwen2-VL’s incorrect answers, where GPT-Score $\leq$ 1. From 1,126 test instances that met this criteria in QaEgo4Dv2, we labeled errors into six key types shown in Table~\ref{tab:error_analysis}. Here are some key takeaways:
\begin{enumerate}
	\item \textbf{Spatial Localization Gaps}: MLLMs struggle with object placement, often predicting generic locations (e.g., ``on the sink'' or ``on the table'') instead of precise ones (e.g., ``inside the square box'' or ``in the drawer'').
	\item \textbf{Object Recognition}: MLLMs struggle to differentiate visually similar objects, misidentifying vegetables (e.g., cabbage as spring onions), liquids (e.g., oil as water), and tools (e.g., wrench as pliers).
	\item \textbf{Video Blur \& Motion}: Low-quality frames hinder MLLMs' ability to distinguish color shades (e.g., predicting ``gray'' instead of ``white'') and object counts.
\end{enumerate}

\begin{table}[h]
\centering
\caption{Error Types and Example Failure Cases}
\begin{tabular}{| m{2.9cm} | m{0.6cm} | m{4cm} |}
\hline
\textbf{Error Type} & \textbf{Freq.} & \textbf{Example Failure Case} \\
\hline
Spatial Localization \newline (Where?) & 57\% & "Where did I put the shoe?" \newline \textbf{Model:} "On the floor" \newline (GT: "At the top of cabinet") \\
\hline
Object Recognition \newline (What?) & 21\% & "What did I pour in the fry pan?" \newline \textbf{Model:} ``Oil'' (GT: ``Soy sauce'') \\
\hline
Color Identification \newline (What color?) & 8\% & "What color was the hanged jacket?" \newline \textbf{Model:} ``black'' (GT: ``dirty white'') \\
\hline
Counting \newline (How many?) & 7\% & "How many pens were on the table?" \newline \textbf{Model:} ``Four'' (GT: ``Seven'') \\
\hline
Person Identification \newline (Who?) & 3\% & "Who did I talk to in the workshop?" \newline \textbf{Model:} ``a man with red shirt'' (GT: ``woman'') \\
\hline
Yes/No Questions \newline (Did I?) & 2\% & ``Did I leave the tap on?'' \newline \textbf{Model:} ``No''  (GT: ``Yes'') \\
\hline
\end{tabular}
\label{tab:error_analysis}
\vspace{-3mm}
\end{table}

\section{Limitations and Future Work}
\textbf{Limitations}:
\begin{enumerate}
	\item \textbf{Residual Noise}: Despite dataset cleaning, some ground-truth intervals or question-answer pairs may remain ambiguous.
	\item \textbf{Computational Cost}: High-resolution frames and large model sizes make training and inference expensive, especially for videos 8+ minutes long.
	\item \textbf{Curriculum Design}: Our static approach to curriculum learning did not yield significant improvements, suggesting a need for more adaptive scheduling or rigorous difficulty metrics.
	\item \textbf{Human vs. Automated Evaluation}: Though we employed LLM-based grading, some subtle queries may require more robust human evaluation.
\end{enumerate}
$\newline$
\textbf{Future Directions}:
\begin{enumerate}
	\item \textbf{Advanced Spatial Reasoning}: Incorporating 3D maps or region-based attention could improve location questions.
	\item \textbf{Dynamic Curriculum}: Explore data-driven methods that rank question difficulty and adapt the training schedule automatically.
	\item \textbf{Longer-Context Architectures}: Efficient attention or memory mechanisms for longer videos may further boost performance.
	\item \textbf{Additional Tasks}: Beyond QA, integrating moment localization, dense captioning, or action forecasting can unlock richer egocentric understanding.
\end{enumerate}

\section{Conclusion and Future Work}
This paper provides a comprehensive study of Multimodal Large Language Models for Egocentric Video Question Answering on QaEgo4Dv2, our refined extension of QaEgo4D. We demonstrate that fine-tuning open-source MLLMs like Video-LLaVa and Qwen2-VL surpass both proprietary models (GPT-4o, Gemini-1.5-Pro) and specialized approaches (GroundVQA variants). Despite notable gains, our error analysis highlights persistent challenges in spatial localization and fine-grained object recognition—critical avenues for future research. By systematically documenting our methods, dataset refinements, and experimental findings, we aim to encourage progress in long-form, real-world egocentric video understanding.

% You must include your signed IEEE copyright release form when you submit your finished paper.
% We MUST have this form before your paper can be published in the proceedings.

% Please direct any questions to the production editor in charge of these proceedings at the IEEE Computer Society Press:
% \url{https://www.computer.org/about/contact}.
%\clearpage
{
    \small
    \bibliographystyle{ieeenat_fullname}
    \bibliography{main}

\begin{thebibliography}{35}
\providecommand{\natexlab}[1]{#1}
\providecommand{\url}[1]{\texttt{#1}}
\expandafter\ifx\csname urlstyle\endcsname\relax
  \providecommand{\doi}[1]{doi: #1}\else
  \providecommand{\doi}{doi: \begingroup \urlstyle{rm}\Url}\fi

\bibitem[hug()]{huggingface}
Hugging face.
\newblock \url{https://huggingface.co/models}.
\newblock Accessed: 2024-12-24.

\bibitem[Bengio et~al.(2009)Bengio, Louradour, Collobert, and Weston]{BengioCurriculum}
Yoshua Bengio, J\'{e}r\^{o}me Louradour, Ronan Collobert, and Jason Weston.
\newblock Curriculum learning.
\newblock In \emph{Proceedings of the 26th Annual International Conference on Machine Learning}, page 41–48, New York, NY, USA, 2009. Association for Computing Machinery.

\bibitem[Bärmann and Waibel(2022)]{qaego4d2022}
Leonard Bärmann and Alex Waibel.
\newblock Where did i leave my keys? — episodic-memory-based question answering on egocentric videos.
\newblock In \emph{2022 IEEE/CVF Conference on Computer Vision and Pattern Recognition Workshops (CVPRW)}, pages 1559--1567, 2022.

\bibitem[Cheng et~al.(2024)Cheng, Guo, Wu, Fang, Li, Liu, and Liu]{Cheng_2024}
Sijie Cheng, Zhicheng Guo, Jinawen Wu, Kechen Fang, Peng Li, Huaping Liu, and Yang Liu.
\newblock Egothink: Evaluating first-person perspective thinking capability of vision-language models.
\newblock In \emph{2024 IEEE/CVF Conference on Computer Vision and Pattern Recognition (CVPR)}, page 14291–14302. IEEE, 2024.

\bibitem[Di and Xie(2024)]{Di_2024}
Shangzhe Di and Weidi Xie.
\newblock Grounded question-answering in long egocentric videos.
\newblock In \emph{2024 IEEE/CVF Conference on Computer Vision and Pattern Recognition (CVPR)}, page 12934–12943. IEEE, 2024.

\bibitem[Fan(2019)]{EgoVQA_Fan}
Chenyou Fan.
\newblock Egovqa - an egocentric video question answering benchmark dataset.
\newblock In \emph{2019 IEEE/CVF International Conference on Computer Vision Workshop (ICCVW)}, pages 4359--4366, 2019.

\bibitem[Fu et~al.(2024)Fu, Dai, Luo, Li, Ren, Zhang, Wang, Zhou, Shen, Zhang, Chen, Li, Lin, Zhao, Li, Xu, Zheng, Chen, Ji, and Sun]{fu2024videommefirstevercomprehensiveevaluation}
Chaoyou Fu, Yuhan Dai, Yongdong Luo, Lei Li, Shuhuai Ren, Renrui Zhang, Zihan Wang, Chenyu Zhou, Yunhang Shen, Mengdan Zhang, Peixian Chen, Yanwei Li, Shaohui Lin, Sirui Zhao, Ke Li, Tong Xu, Xiawu Zheng, Enhong Chen, Rongrong Ji, and Xing Sun.
\newblock Video-mme: The first-ever comprehensive evaluation benchmark of multi-modal llms in video analysis, 2024.

\bibitem[Grauman et~al.(2022)Grauman, Westbury, Byrne, Chavis, Furnari, Girdhar, Hamburger, Jiang, Liu, Liu, Martin, Nagarajan, Radosavovic, Ramakrishnan, Ryan, Sharma, Wray, Xu, Xu, Zhao, Bansal, Batra, Cartillier, Crane, Do, Doulaty, Erapalli, Feichtenhofer, Fragomeni, Fu, Gebreselasie, Gonzalez, Hillis, Huang, Huang, Jia, Khoo, Kolar, Kottur, Kumar, Landini, Li, Li, Li, Mangalam, Modhugu, Munro, Murrell, Nishiyasu, Price, Puentes, Ramazanova, Sari, Somasundaram, Southerland, Sugano, Tao, Vo, Wang, Wu, Yagi, Zhao, Zhu, Arbelaez, Crandall, Damen, Farinella, Fuegen, Ghanem, Ithapu, Jawahar, Joo, Kitani, Li, Newcombe, Oliva, Park, Rehg, Sato, Shi, Shou, Torralba, Torresani, Yan, and Malik]{Grauman_2022}
Kristen Grauman, Andrew Westbury, Eugene Byrne, Zachary Chavis, Antonino Furnari, Rohit Girdhar, Jackson Hamburger, Hao Jiang, Miao Liu, Xingyu Liu, Miguel Martin, Tushar Nagarajan, Ilija Radosavovic, Santhosh~Kumar Ramakrishnan, Fiona Ryan, Jayant Sharma, Michael Wray, Mengmeng Xu, Eric~Zhongcong Xu, Chen Zhao, Siddhant Bansal, Dhruv Batra, Vincent Cartillier, Sean Crane, Tien Do, Morrie Doulaty, Akshay Erapalli, Christoph Feichtenhofer, Adriano Fragomeni, Qichen Fu, Abrham Gebreselasie, Cristina Gonzalez, James Hillis, Xuhua Huang, Yifei Huang, Wenqi Jia, Weslie Khoo, Jachym Kolar, Satwik Kottur, Anurag Kumar, Federico Landini, Chao Li, Yanghao Li, Zhenqiang Li, Karttikeya Mangalam, Raghava Modhugu, Jonathan Munro, Tullie Murrell, Takumi Nishiyasu, Will Price, Paola~Ruiz Puentes, Merey Ramazanova, Leda Sari, Kiran Somasundaram, Audrey Southerland, Yusuke Sugano, Ruijie Tao, Minh Vo, Yuchen Wang, Xindi Wu, Takuma Yagi, Ziwei Zhao, Yunyi Zhu, Pablo Arbelaez, David Crandall, Dima Damen, Giovanni~Maria
  Farinella, Christian Fuegen, Bernard Ghanem, Vamsi~Krishna Ithapu, C.~V. Jawahar, Hanbyul Joo, Kris Kitani, Haizhou Li, Richard Newcombe, Aude Oliva, Hyun~Soo Park, James~M. Rehg, Yoichi Sato, Jianbo Shi, Mike~Zheng Shou, Antonio Torralba, Lorenzo Torresani, Mingfei Yan, and Jitendra Malik.
\newblock Ego4d: Around the world in 3,000 hours of egocentric video.
\newblock In \emph{2022 IEEE/CVF Conference on Computer Vision and Pattern Recognition (CVPR)}, page 18973–18990. IEEE, 2022.

\bibitem[Jang et~al.(2017)Jang, Song, Yu, Kim, and Kim]{Jang_2017}
Yunseok Jang, Yale Song, Youngjae Yu, Youngjin Kim, and Gunhee Kim.
\newblock Tgif-qa: Toward spatio-temporal reasoning in visual question answering.
\newblock In \emph{2017 IEEE Conference on Computer Vision and Pattern Recognition (CVPR)}, page 1359–1367. IEEE, 2017.

\bibitem[Jia et~al.(2022)Jia, Lei, Zhu, and Huang]{jia2022egotaskqa}
Baoxiong Jia, Ting Lei, Song-Chun Zhu, and Siyuan Huang.
\newblock Egotask{QA}: Understanding human tasks in egocentric videos.
\newblock In \emph{Thirty-sixth Conference on Neural Information Processing Systems Datasets and Benchmarks Track}, 2022.

\bibitem[Li et~al.(2022{\natexlab{a}})Li, Niu, and Zhang]{Li_2022}
Jiangtong Li, Li Niu, and Liqing Zhang.
\newblock From representation to reasoning: Towards both evidence and commonsense reasoning for video question-answering.
\newblock In \emph{2022 IEEE/CVF Conference on Computer Vision and Pattern Recognition (CVPR)}, page 21241–21250. IEEE, 2022{\natexlab{a}}.

\bibitem[Li et~al.(2020)Li, Chen, Cheng, Gan, Yu, and Liu]{Li_2020}
Linjie Li, Yen-Chun Chen, Yu Cheng, Zhe Gan, Licheng Yu, and Jingjing Liu.
\newblock Hero: Hierarchical encoder for video+language omni-representation pre-training.
\newblock In \emph{Proceedings of the 2020 Conference on Empirical Methods in Natural Language Processing (EMNLP)}. Association for Computational Linguistics, 2020.

\bibitem[Li et~al.(2022{\natexlab{b}})Li, Yang, and Zou]{9737531}
Shanhao Li, Bang Yang, and Yuexian Zou.
\newblock Adaptive curriculum learning for video captioning.
\newblock \emph{IEEE Access}, 10:\penalty0 31751--31759, 2022{\natexlab{b}}.

\bibitem[Li et~al.(2023)Li, Wang, and Jia]{li2023llamavidimageworth2}
Yanwei Li, Chengyao Wang, and Jiaya Jia.
\newblock Llama-vid: An image is worth 2 tokens in large language models, 2023.

\bibitem[Lin et~al.(2023)Lin, Zhu, Ye, Ning, Jin, and Yuan]{lin2023video}
Bin Lin, Bin Zhu, Yang Ye, Munan Ning, Peng Jin, and Li Yuan.
\newblock Video-llava: Learning united visual representation by alignment before projection.
\newblock \emph{arXiv preprint arXiv:2311.10122}, 2023.

\bibitem[Liu et~al.(2024)Liu, Li, Li, Li, Zhang, Shen, and Lee]{liu2024llavanext}
Haotian Liu, Chunyuan Li, Yuheng Li, Bo Li, Yuanhan Zhang, Sheng Shen, and Yong~Jae Lee.
\newblock Llava-next: Improved reasoning, ocr, and world knowledge, 2024.

\bibitem[Majumdar et~al.(2024)Majumdar, Ajay, Zhang, Putta, Yenamandra, Henaff, Silwal, Mcvay, Maksymets, Arnaud, Yadav, Li, Newman, Sharma, Berges, Zhang, Agrawal, Bisk, Batra, Kalakrishnan, Meier, Paxton, Sax, and Rajeswaran]{OpenEQA2023}
Arjun Majumdar, Anurag Ajay, Xiaohan Zhang, Pranav Putta, Sriram Yenamandra, Mikael Henaff, Sneha Silwal, Paul Mcvay, Oleksandr Maksymets, Sergio Arnaud, Karmesh Yadav, Qiyang Li, Ben Newman, Mohit Sharma, Vincent Berges, Shiqi Zhang, Pulkit Agrawal, Yonatan Bisk, Dhruv Batra, Mrinal Kalakrishnan, Franziska Meier, Chris Paxton, Sasha Sax, and Aravind Rajeswaran.
\newblock Openeqa: Embodied question answering in the era of foundation models.
\newblock In \emph{Conference on Computer Vision and Pattern Recognition (CVPR)}, 2024.

\bibitem[Mangalam et~al.(2023)Mangalam, Akshulakov, and Malik]{mangalam2023egoschemadiagnosticbenchmarklongform}
Karttikeya Mangalam, Raiymbek Akshulakov, and Jitendra Malik.
\newblock Egoschema: A diagnostic benchmark for very long-form video language understanding, 2023.

\bibitem[OpenAI(2024)]{openai2024gpt4o}
OpenAI.
\newblock Hello gpt-4o.
\newblock \url{https://openai.com/index/hello-gpt-4o/⁠}, 2024.
\newblock Accessed: 2024-12-24.

\bibitem[Reimers and Gurevych(2019)]{Reimers2019SentenceBERTSE}
Nils Reimers and Iryna Gurevych.
\newblock Sentence-bert: Sentence embeddings using siamese bert-networks.
\newblock In \emph{Conference on Empirical Methods in Natural Language Processing}, 2019.

\bibitem[Team(2024)]{gemini1_5Pro2024}
Gemini Team.
\newblock Gemini 1.5: Unlocking multimodal understanding across millions of tokens of context.
\newblock \url{https://storage.googleapis.com/deepmind-media/gemini/gemini_v1_5_report.pdf}, 2024.
\newblock Accessed: 2024-12-24.

\bibitem[Wang et~al.(2024)Wang, Bai, Tan, Wang, Fan, Bai, Chen, Liu, Wang, Ge, Fan, Dang, Du, Ren, Men, Liu, Zhou, Zhou, and Lin]{wang2024qwen2vl}
Peng Wang, Shuai Bai, Sinan Tan, Shijie Wang, Zhihao Fan, Jinze Bai, Keqin Chen, Xuejing Liu, Jialin Wang, Wenbin Ge, Yang Fan, Kai Dang, Mengfei Du, Xuancheng Ren, Rui Men, Dayiheng Liu, Chang Zhou, Jingren Zhou, and Junyang Lin.
\newblock Qwen2-vl: Enhancing vision-language model's perception of the world at any resolution, 2024.

\bibitem[Wu et~al.(2024)Wu, Li, Chen, and Li]{wu2024longvideobenchbenchmarklongcontextinterleaved}
Haoning Wu, Dongxu Li, Bei Chen, and Junnan Li.
\newblock Longvideobench: A benchmark for long-context interleaved video-language understanding, 2024.

\bibitem[Xiao et~al.(2021)Xiao, Shang, Yao, and Chua]{Xiao_2021}
Junbin Xiao, Xindi Shang, Angela Yao, and Tat-Seng Chua.
\newblock Next-qa: Next phase of question-answering to explaining temporal actions.
\newblock In \emph{2021 IEEE/CVF Conference on Computer Vision and Pattern Recognition (CVPR)}. IEEE, 2021.

\bibitem[Xu et~al.(2017)Xu, Zhao, Xiao, Wu, Zhang, He, and Zhuang]{MSVDQA}
Dejing Xu, Zhou Zhao, Jun Xiao, Fei Wu, Hanwang Zhang, Xiangnan He, and Yueting Zhuang.
\newblock Video question answering via gradually refined attention over appearance and motion.
\newblock In \emph{Proceedings of the 25th ACM International Conference on Multimedia}, page 1645–1653, New York, NY, USA, 2017. Association for Computing Machinery.

\bibitem[Xu et~al.(2023)Xu, Lan, Xie, Chen, and Lu]{xu2023retrievalbasedvideolanguagemodel}
Jiaqi Xu, Cuiling Lan, Wenxuan Xie, Xuejin Chen, and Yan Lu.
\newblock Retrieval-based video language model for efficient long video question answering, 2023.

\bibitem[Xu et~al.(2024)Xu, Zhao, Zhou, Lin, Ng, and Feng]{xu2024pllavaparameterfreellava}
Lin Xu, Yilin Zhao, Daquan Zhou, Zhijie Lin, See~Kiong Ng, and Jiashi Feng.
\newblock Pllava : Parameter-free llava extension from images to videos for video dense captioning, 2024.

\bibitem[Ye et~al.(2024)Ye, Zhang, Daxberger, Chen, Lin, Li, Zhang, You, Xu, Gan, Lu, and Yang]{ye2024mmegobuildingegocentricmultimodal}
Hanrong Ye, Haotian Zhang, Erik Daxberger, Lin Chen, Zongyu Lin, Yanghao Li, Bowen Zhang, Haoxuan You, Dan Xu, Zhe Gan, Jiasen Lu, and Yinfei Yang.
\newblock Mm-ego: Towards building egocentric multimodal llms, 2024.

\bibitem[Yu et~al.(2019)Yu, Xu, Yu, Yu, Zhao, Zhuang, and Tao]{Yu_2019}
Zhou Yu, Dejing Xu, Jun Yu, Ting Yu, Zhou Zhao, Yueting Zhuang, and Dacheng Tao.
\newblock Activitynet-qa: A dataset for understanding complex web videos via question answering.
\newblock \emph{Proceedings of the AAAI Conference on Artificial Intelligence}, 33\penalty0 (01):\penalty0 9127–9134, 2019.

\bibitem[Zhang et~al.(2023)Zhang, Li, and Bing]{zhang2023videollama}
Hang Zhang, Xin Li, and Lidong Bing.
\newblock Video-llama: An instruction-tuned audio-visual language model for video understanding, 2023.

\bibitem[Zhang et~al.(2024{\natexlab{a}})Zhang, Zhang, Li, Zeng, Yang, Zhang, Wang, Tan, Li, and Liu]{zhang2024longcontexttransferlanguage}
Peiyuan Zhang, Kaichen Zhang, Bo Li, Guangtao Zeng, Jingkang Yang, Yuanhan Zhang, Ziyue Wang, Haoran Tan, Chunyuan Li, and Ziwei Liu.
\newblock Long context transfer from language to vision, 2024{\natexlab{a}}.

\bibitem[Zhang et~al.(2024{\natexlab{b}})Zhang, Li, Liu, Lee, Gui, Fu, Feng, Liu, and Li]{zhang2024llavanextvideo}
Yuanhan Zhang, Bo Li, haotian Liu, Yong~jae Lee, Liangke Gui, Di Fu, Jiashi Feng, Ziwei Liu, and Chunyuan Li.
\newblock Llava-next: A strong zero-shot video understanding model, 2024{\natexlab{b}}.

\bibitem[Zheng et~al.(2023)Zheng, Chiang, Sheng, Zhuang, Wu, Zhuang, Lin, Li, Li, Xing, Zhang, Gonzalez, and Stoica]{Zheng2023JudgingLW}
Lianmin Zheng, Wei-Lin Chiang, Ying Sheng, Siyuan Zhuang, Zhanghao Wu, Yonghao Zhuang, Zi Lin, Zhuohan Li, Dacheng Li, Eric~P. Xing, Haotong Zhang, Joseph~E. Gonzalez, and Ion Stoica.
\newblock Judging llm-as-a-judge with mt-bench and chatbot arena.
\newblock \emph{ArXiv}, abs/2306.05685, 2023.

\bibitem[Zhou et~al.(2024)Zhou, Shu, Zhao, Wu, Xiao, Yang, Xiong, Zhang, Huang, and Liu]{zhou2024mlvucomprehensivebenchmarkmultitask}
Junjie Zhou, Yan Shu, Bo Zhao, Boya Wu, Shitao Xiao, Xi Yang, Yongping Xiong, Bo Zhang, Tiejun Huang, and Zheng Liu.
\newblock Mlvu: A comprehensive benchmark for multi-task long video understanding, 2024.

\bibitem[Zhu et~al.(2023)Zhu, Lin, Ning, Yan, Cui, Wang, Pang, Jiang, Zhang, Li, et~al.]{zhu2023languagebind}
Bin Zhu, Bin Lin, Munan Ning, Yang Yan, Jiaxi Cui, HongFa Wang, Yatian Pang, Wenhao Jiang, Junwu Zhang, Zongwei Li, et~al.
\newblock Languagebind: Extending video-language pretraining to n-modality by language-based semantic alignment.
\newblock \emph{arXiv preprint arXiv:2310.01852}, 2023.

\end{thebibliography}
}

% WARNING: do not forget to delete the supplementary pages from your submission 
\clearpage
\setcounter{page}{1}
\maketitlesupplementary

% \section{Rationale}
% \label{sec:rationale}
\appendix
\section{Examples of noisy instances in QaEgo4D dataset}
\label{sec:appendix_d}
\begin{table}[ht]
\centering
\caption{Examples of non-coherent answers for question types that were removed from QaEgo4D dataset}
\label{tab:dataset}
\begin{tabular}{ |p{5cm}|p{2cm}| } 
\hline
\textbf{Question} & \textbf{Answer}  \\
\hline
Who did I talk to when I was standing in the garage? & 2  \\ \hline
Who did I talk to at the garage? & 1 \\ \hline
what did I put in the bucket? & on the counter  \\ \hline
what did I put in the kitchen cabinet? & rinses  \\ \hline
Where was my remote control in the living room? & yes \\ \hline
Where is the broom leaned on the wall? & no\\
\hline
\end{tabular}
\end{table}

\section{Prompts used for evaluation of pre-trained models}
\label{sec:appendix_a}
\subsection{OpenQA}
\textbf{For GPT-4o, Gemini-1.5-Pro, and Qwen2-VL Evaluation:}\\
Answer the question based on the given video.\\
Answer should be concise phrase. \\
Question: \{question\}\\
Answer: 
\\
\\
\textbf{For Video-LLaVa Evaluation:}\\
USER: ⟨video⟩ Question: \{question\} Please give short phrase in answer. \\
ASSISTANT: Answer:

\subsection{CloseQA}
\textbf{For GPT-4o, Gemini-1.5-Pro, and Qwen2-VL Evaluation:}
\\
Select the correct answer strictly from the provided multiple-choice options based on the video content. Do not generate any extra text or explanation—just output the selected answer exactly as written in the choices.\\
\\
Question: \{question\}\\
Multiple-choice options:\\\{multiple\_choice\}\\
Answer:\\
\textbf{For Video-LLaVa Evaluation:}\\
USER: ⟨video⟩
{question}
Pick the best answer from the following multiple choices.\\
\{multiple\_choice\}\\
ASSISTANT:

\section{Prompts used for LLM-as-Judge Metrics}
\label{sec:appendix_b}
\textbf{For GPT4o as Judge}
\\
You are an intelligent chatbot designed for evaluating the correctness of generative outputs for question-answer pairs.\\
You are tasked with evaluating how well the predicted answers match the ground truth answers. Focus on the meaning and intent of the answer rather than exact wording. A match should be considered acceptable if:\\
\\
Semantic Equivalence: The predicted answer conveys the same overall meaning as the ground truth answer, even if the wording or phrasing is different.\\
\\
Contextual Match: Minor differences in description or detail (e.g., specific object characteristics like size, position, or color) should be considered a match if they still refer to the same object or event in context.\\
\\
Synonyms and Paraphrases: If the predicted answer uses a synonym, paraphrase, or a more general/specific term than the ground truth, it can be considered a match as long as it describes the same concept or entity.\\
\\
Partial Matches: If the predicted answer captures a significant part of the ground truth but omits or changes some details, it may still be considered a match if the core meaning is preserved in context of question.\\
\\
Examples:\\
``one'' and ``one large spoon'': Consider these a match if ``one'' captures the essence of quantity.\\
``on the floor'' and ``under the table'': Although slightly different, if they refer to similar spatial positioning in context, consider them a match.\\
``blue'' and ``blue and black'': If ``blue'' is a dominant color, consider them a match.
``vegetable'' and ``okra'': Consider these a match as ``okra'' is essentially a vegetable\\
\\
Please evaluate the following image-based question-answer pair:\\
Question: \{question\}\\
Correct Answer: \{answer\}\\
Predicted Answer: \{prediction\}\\
\\
Provide your evaluation only as a yes/no and score where the score is an integer value between 0 and 5, with 5 indicating the highest meaningful match.\\
Please generate the response in the form of a Python dictionary string with keys 'pred' and `score', where value of `pred' is a string of `yes' or `no' and value of `score' is in INTEGER, not STRING.\\
\\
DO NOT PROVIDE ANY OTHER OUTPUT TEXT OR EXPLANATION. Only provide a syntactically valid Python dictionary string.\\
For example, your response should look like this: \{``pred'': ``yes'', ``score'': 4\}.\\
\\
\textbf{For Gemini-1.5-flash as Judge}
\\
You are an intelligent chatbot designed for evaluating the correctness of generative outputs for question-answer pairs for a given video.\\
Your task is to compare the predicted answer with the correct answer and determine if they match meaningfully. Here's how you can accomplish the task:\\
------\\
\\
\#\#INSTRUCTIONS:\\
- Consider the content of the video.\\
- Focus on the meaningful match between the predicted answer and the correct answer.\\
- Consider synonyms or paraphrases as valid matches.\\
- Evaluate the correctness of the prediction compared to the answer.\\
\\
Please evaluate the following video-based question-answer pair:\\
Question: \{question\} \\
Correct Answer: \{answer\} \\
Predicted Answer: \{prediction\} \\
\\
Provide your evaluation only as a yes/no and score where the score is an integer value between 0 and 5, with 5 indicating the highest meaningful match.\\
\\
Please generate the response in the form of a Python dictionary string with keys `pred' and `score', where value of `pred' is a string of `yes' or `no', indicating predicted answer match with correct answer and value of `score' is in INTEGER, not STRING.\\
\\
DO NOT PROVIDE ANY OTHER OUTPUT TEXT OR EXPLANATION. Only provide a syntactically valid Python dictionary string.\\
For example, your response should look like this: \{``pred'': ``yes'', ``score'': 4\}.

\section{Measuring effect of image resolution}
\label{sec:appendix_c}
\begin{table}[ht]
\centering
\caption{Experiments on different frame resolution on QaEgo4Dv2 test set. Fixed number of frames 16.}
\label{tab:imresolution}
\begin{tabular}{ |p{2.5cm}|p{2cm}|p{2cm}|} 
\hline
\textbf{Model} & \textbf{Resolution} & \textbf{Sim}  \\
\hline
\multirow{3}{=}{GPT4o} & 224x & 54.83 \\ 
 & \textbf{336x} & \textbf{55.06} \\ 
 & 512x &  54.36\\ \hline
\end{tabular}
\end{table}

\end{document}